# Dynamic programming in influence diagrams with decision circuits

**Ross D. Shachter**
Management Science and Engineering
Stanford University
Stanford, CA 94305, USA
shachter@stanford.edu

**Debarun Bhattacharjya**
Business Analytics and Math Sciences
IBM T. J. Watson Research Center
Yorktown Heights, NY 10598, USA
debarunb@us.ibm.com

## Abstract

Decision circuits perform efficient evaluation of influence diagrams, building on the advances in arithmetic circuits for belief network inference [Darwiche, 2003; Bhattacharjya and Shachter, 2007]. We show how even more compact decision circuits can be constructed for dynamic programming in influence diagrams with separable value functions and conditionally independent subproblems. Once a decision circuit has been constructed based on the diagram's "global" graphical structure, it can be compiled to exploit "local" structure for efficient evaluation and sensitivity analysis.

## 1 INTRODUCTION

Influence diagrams [Howard and Matheson, 1984] are popular graphical models for representing, evaluating and analyzing a single rational decision maker's decision problem. We can exploit any separable value functions in influence diagrams for more efficient evaluation [Tatman and Shachter, 1990; Luque and Diez, 2009]. Decision circuits are graphical representations for efficient evaluation and sensitivity analysis on influence diagrams with a single value node [Bhattacharjya and Shachter, 2007, 2008, 2010]. In this paper, we show how they can be applied to influence diagrams with multiple value functions using join tree methods [Shachter and Peot, 1992; Dittmer and Jensen, 1994].

Arithmetic circuits [Darwiche, 2003] are graphical representations for probabilistic inference and sensitivity analysis in Bayesian belief networks. Decision circuits are a natural extension to arithmetic circuits for evaluating sequential decision problems represented as influence diagrams.

Global structure refers to the topology of the graph, the size and the treewidth for a given task. Techniques have been developed to exploit global structure in the construction of arithmetic circuits applying logic [Darwiche, 2002; Chavira and Darwiche, 2005; Chavira, 2007] and join trees [Park and Darwiche, 2004]. We apply join tree methods, as well, but the arithmetic circuit approaches do not deal with the additional vertex elimination order restrictions in decision problems and complex accounting for separable value functions. To address these issues, we introduce a "branching operator" for decision circuits, use directed chordal graphs to develop our join tree structure, and introduce decision circuit backbones as a compact representation for symmetric decision circuits.

Local structure in belief networks refers to the specific numbers, particularly zeros, in the conditional probability tables (CPTs). Like arithmetic circuits, decision circuits are compiled so that subsequent evaluation and analysis can be more efficient. This is particularly useful for real-time decision making.

This paper is organized as follows. In Section 2 we review influence diagrams and introduce some examples. In Section 3 we review decision circuits, and introduce a new type of decision circuit operator for branching. In Section 4 we show how to construct a decision circuit for an influence diagram with multiple value nodes, based on directed chordal graphs and introducing an intermediate structure we call a "decision circuit backbone." In Section 5 we discuss methods to compile the decision circuit, taking advantage of local structure. Finally, we conclude the paper in Section 6 with some suggestions for future research.

## 2 INFLUENCE DIAGRAMS

In this section we review influence diagrams and introduce the notation and two examples that we will refer to throughout the paper.

An *influence diagram* is a directed acyclic graphical model. The nodes correspond to the variables, uncertainties drawn as ovals, decisions drawn as rectangles, and values drawn as rounded rectangles. We denote variables by upper-case letters ($X$) and their possible states by lower-case letters ($x$). A bold-faced letter (**U**) indicates a set of variables. We refer interchangeably to a node and its corresponding variable, referring to them simply as uncertainties, decisions, and values. If $X$ is a node with parents **U**, then $X$**U** is called the *family* of $X$. The parents of an uncertain or value node condition its probability distribution or expected value, while the parents of a decision node will be observed before the decision will be made.

Consider the influence diagram shown in Figure 1a from Dittmer and Jensen [1997]. There are three decisions, $D1$, $D2$, and $D3$, one value, $V$, and four uncertainties, $A$, $B$, $C$, and $E$. The distribution for $B$ is unconditional, while all of the other distributions are conditional. No other variables will be observed before decision $D1$ is made, but when $D3$ is made $D1$, C, $D2$, $E$, and $A$ will have been observed.

Uncertain descendants of decisions are said to be *responsive* because their outcome can be affected by the decisions. The only non-responsive uncertainty is $B$, and thus it is the only uncertainty which could be observed in advance [Heckerman and Shachter, 1995]. Such evidence can incorporate any data already collected or determine the value of clairvoyance. We also allow evidence to indicate which alternatives are available and which value variables are to be accumulated.

We assume that value nodes have no children and that when there are multiple values the total value is their sum [Tatman and Shachter, 1990]. (The methods in this paper could also be applied to multilinear value models with nested sums and products.) Consider the influence diagram shown in Figure 2a from Jensen et al [1994]. There are four decisions in this example and four values, and the decisions are made so as to maximize the sum of the values. There are three non-responsive uncertainties, $A$, $B$, and $C$.

In addition to the structure of the influence diagram, we need the possible states for each of the uncertainties and the decisions. For each uncertainty $X$ with parents **U** there is a distribution $\theta_{x|\mathbf{u}} = P\{X = x|\mathbf{U} = \mathbf{u}\}$. For each value $V$ with parents **U**, we treat $v$ as a placeholder in $\theta_{v|\mathbf{u}} = E\{V|\mathbf{U} = \mathbf{u}\}$, which we assume is positive without any loss of generality. (As a result, if one of the alternatives for a decision is unavailable it will yield zero value and be dominated by any available alternative.)

We allow evidence **e** about uncertain variables **E**, and augmented evidence **e**′, which also includes the value distributions, $\{\theta_{x|\mathbf{u}}\}$, and any evidence about values and decisions. For any uncertainty or decision $X$, $\lambda_x = 0$ if we assert that $X \neq x$ and $\lambda_x = 1$ otherwise. In the case of value variable $V$, we allow any $\lambda_v \geq 0$ to weight or discount its contribution, with $\lambda_v = 0$ if we want to ignore the value variable $V$ in making our decisions.

We can recognize which of the available observations are *requisite* for each of the decisions. In the example shown in Figure 1a, even though $D1$, $C$, $D2$, $E$, and $A$ are all known at the time of decision $D3$, $A$ is sufficient to make the best decision. This can be determined for each decision, in reverse order, by finding the observations relevant to the value descendants of the decision [Shachter, 1998, 1999].

We assume that the influence diagram is a single component. If not, the problem can be solved separately for each component. We also assume that there are no nodes in the diagram which could be simply removed. For example, some non-responsive uncertainties might become relevant if they were observed.

For each of the diagrams, a *moral graph*, similar to the ones in Shachter [1999], is shown in Figures 1b and 2b. Moral graphs are obtained by including only the requisite observations as parents of the decisions and adding undirected edges (as dashed lines), if necessary, between any two nodes with a common child. These extra edges will ensure that each node's family can be represented in the constructed decision circuit, so that the distribution can be incorporated.

The standard exact approaches to evaluating influence diagrams incorporate an implicit or explicit *variable elimination order*. If requisite observations for a decision precede the decision while responsive uncertain and value variables follow it in the order, the order is said to be *consistent* with the influence diagram. In section 4 we will construct a decision circuit based on any given *target* variable elimination order consistent with the influence diagram.

## 3 DECISION CIRCUITS

Decision circuits are modified arithmetic circuits [Darwiche, 2003] for efficient evaluation and analysis of influence diagrams. In particular, they have a maximization operator in addition to sum and product operators [Bhattacharjya and Shachter, 2007], and they perform evaluation with an upward sweep through the circuit and differentiation with a subsequent downward sweep. In this paper, we also introduce a branching operator to deal with conditionally independent subproblems and additive value functions. The branching operator requires that we evaluate two functions simultaneously as we sweep upward through the decision circuit, one for the expected value weighted by the probability of the evidence and another with only the probability of the evidence. On the downward pass, we need to evaluate three sets of partial derivatives.

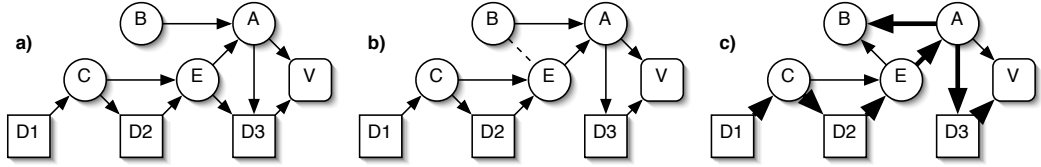

Figure 1: Original influence diagram, moral graph, and directed chordal graph for the Dittmer and Jensen [1997] example with elimination order D1, C, D2, E, A, B, D3, V.

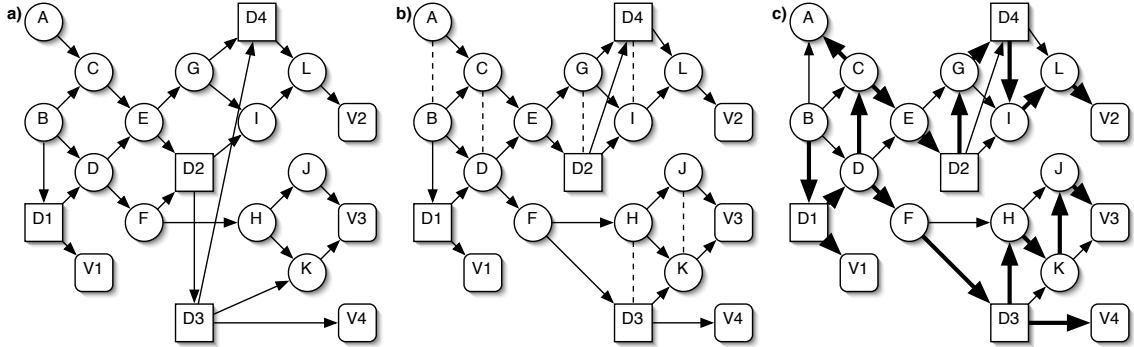

Figure 2: Original influence diagram, moral graph, and directed chordal graph for the Jensen et al [1994] example with elimination order B, D1, D, C, A, E, D2, G, D4, I, L, F, D3, H, K, J, V1, V2, V3, V4.

A sequential decision problem can be represented as a dynamic programming function $g(\mathbf{e}')$ computing the maximal expected value weighted by the probability of the evidence about any of the variables, including the value variables. Similarly, we can represent the probability of the evidence as $g(\mathbf{e})$, where evidence $\mathbf{e}$ is the evidence about just the uncertain variables, so the maximal expected value is given by $g(\mathbf{e}')/g(\mathbf{e})$. The function $g(\mathbf{e}')$ is our objective for making decisions and $g(\mathbf{e})$ normalizes it to units of expected value.

We derive in the next section the functions below for the influence diagram shown in Figure 1a:

$$g(\mathbf{e}) = \sum_c \lambda_c \theta_{c|d1} \sum_e \lambda_e \theta_{e|c,d2} \sum_a \sum_b \lambda_a \theta_{a|b,e} \lambda_b \theta_b$$
$$g(\mathbf{e}') = \max_{d1} \lambda_{d1} \sum_c \lambda_c \theta_{c|d1} \max_{d2} \lambda_{d2} \sum_e \lambda_e \theta_{e|c,d2}$$
$$\sum_a [\sum_b \lambda_a \theta_{a|b,e} \lambda_b \theta_b][\max_{d3} \lambda_{d3} \lambda_v \theta_{v|a,d3}].$$

A *decision circuit* is a graphical representation of a dynamic programming function. It is a directed acyclic graph with leaves 0, 1, $\{\theta_{x|\mathbf{u}}\}$, and $\{\lambda_x\}$. The non-leaf nodes are operators, sum, product, max, and branching, on the node's children. There is a single root node, corresponding to our functions, $g(\mathbf{e}')$ and $g(\mathbf{e})$. The *size* of a decision circuit is the number of arcs it contains. We have not displayed the decision circuit for the diagram in Figure 1a because it is too large. Even if each variable had only two possibilities, the circuit would have 122 nodes and 138 arcs. Although a decision circuit is an efficient representation for analyzing the problem, it is not well suited for communication, and therefore we introduce decision circuit backbones in the next section as a more compact and readable intermediate representation.

The decision circuit is not a tree, because nodes can have multiple parents to exploit conditional independence [Darwiche, 2000]. For example, in the computation of $g$ above, $c$ and $d2$ play no role after $\theta_{e|c,d2}$ is introduced, and the rest of the expression can be computed for any possible $(c, d2)$. This would be implemented in a decision circuit by having all possible $(c, d2)$ as parents of the $A$ sum node for a particular $e$, an example of coalescence, discussed in Section 5.

The functions $g(\mathbf{e})$ and $g(\mathbf{e}')$ are computed together by *evaluating* the circuit in an *upward pass*, starting just above the leaves and ending at the root, only visiting a node after all of its (non-leaf) children have been visited and then performing the node's operation. For each intermediate node $x$ there are two values, $g_x(\mathbf{e}_x)$ and $g_x(\mathbf{e}'_x)$, where $\mathbf{e}_x$ represents the evidence in the leaves below $x$. For each of the functions, the value of a sum node is the sum of the corresponding values of its children. The value of a product node is their product, although to compute $g_x(\mathbf{e}_x)$ we ignore any $\lambda_d$, $\lambda_v$, and $\theta_{v|\mathbf{u}}$ children. (The lists of which children contribute to both $g_x(\mathbf{e}_x)$ and $g_x(\mathbf{e}'_x)$ and which contribute only

to $g_x(\mathbf{e}'_x)$ can be created at compile time.)

The $g_x(\mathbf{e}'_x)$ value of a max node $x$ is the largest corresponding value of its children, remembering which child $y$ yielded the maximum, and breaking ties arbitrarily. The $g_x(\mathbf{e}_x)$ value of a max node $x$ can be taken from any of its children when $\mathbf{e}$ includes only non-responsive evidence, since they should all be equal, and we therefore suggest using $g_y(\mathbf{e}_y)$.

### 3.1 A New Branching Operator

We add a *branching node* to the decision circuit to combine conditionally independent subproblems with their own additive value functions and disjoint evidence. The branching node $x$ works differently for the two functions: $g_x(\mathbf{e}_x) = g_1(\mathbf{e}_1)g_2(\mathbf{e}_2)$ and $g_x(\mathbf{e}'_x) = g_1(\mathbf{e}'_1)g_2(\mathbf{e}_2) + g_1(\mathbf{e}_1)g_2(\mathbf{e}'_2)$, the cross-product of the two child branches [Shachter and Peot, 1992; Dittmer and Jensen, 1994]. This distinction is necessary to normalize the different values of $g_i(\mathbf{e}')$ before summing, so that the combined $g_x(\mathbf{e}'_x)$ function incorporates all evidence from both branch 1 and branch 2. (This can be generalized to more than two branches, but it is more efficient to create a binary tree of branching nodes.)

The partial derivatives of $g(\mathbf{e})$ and $g(\mathbf{e}')$ can then be computed by *differentiating* the circuit in a *downward pass*, starting at the root node and ending at the leaves, only visiting a node after all of its parents have been visited. For each node $x$ there are three partial derivatives: $\frac{\partial g(\mathbf{e})}{\partial g_x(\mathbf{e}_x)}$, $\frac{\partial g(\mathbf{e}')}{\partial g_x(\mathbf{e}_x)}$, and $\frac{\partial g(\mathbf{e}')}{\partial g_x(\mathbf{e}'_x)}$. There are three derivatives because $g(\mathbf{e})$ can be computed solely from the local probability functions as in an arithmetic circuit, but we use both local functions to compute $g(\mathbf{e}')$ with the branching operator.

For the downward differentiation pass, because the root is not a branching node we can initialize the root derivatives to $\frac{\partial g(\mathbf{e})}{\partial g(\mathbf{e})} = 1$, $\frac{\partial g(\mathbf{e}')}{\partial g(\mathbf{e})} = 0$, and $\frac{\partial g(\mathbf{e}')}{\partial g(\mathbf{e}')} = 1$. Each of the derivatives at all other nodes are sums over their parents, applying the chain rule. Each child of a sum node inherits its parent's corresponding derivatives; each child of a product node inherits its parent's corresponding derivatives multiplied by its siblings' corresponding values; and the maximizing child of a max node inherits its parent's corresponding derivatives. The first child of a branching node $x$ inherits $g_2(\mathbf{e}_2)\frac{\partial g(\mathbf{e})}{\partial g_x(\mathbf{e}_x)}$ toward $\frac{\partial g(\mathbf{e})}{\partial g_1(\mathbf{e}_x)}$, $g_2(\mathbf{e}_2)\frac{\partial g(\mathbf{e}')}{\partial g_x(\mathbf{e}_x)} + g_2(\mathbf{e}'_2)\frac{\partial g(\mathbf{e}')}{\partial g_x(\mathbf{e}'_x)}$ toward $\frac{\partial g(\mathbf{e}')}{\partial g_1(\mathbf{e}_x)}$, and $g_2(\mathbf{e}_2)\frac{\partial g(\mathbf{e}')}{\partial g_x(\mathbf{e}'_x)}$ toward $\frac{\partial g(\mathbf{e}')}{\partial g_1(\mathbf{e}'_x)}$, with symmetric expressions for the second child.

The upward and downward passes are also referred to as *sweeps*. Thus we can evaluate the decision problem with one sweep and compute all of the derivatives with one another. Consequently the time complexity for these operations is linear in the size of the circuit.

## 4 BUILDING DECISION CIRCUITS FROM INFLUENCE DIAGRAMS

In this section we show how to build a decision circuit that exploits the global structure of an influence diagram, given a target variable elimination order. After constructing a directed chordal graph from the moral graph using the target order, we build a "decision circuit backbone" to specify the decision circuit.

### 4.1 Directed chordal graphs

The theory of chordal graphs and junction trees is well developed and we build on these results [Golumbic, 1980; Tarjan and Yannakakis, 1984; Shachter et al, 1990]. We define an acyclic directed graph to be *chordal* if there is an arc between any two nodes with a common child. A directed chordal graph can always be obtained from an undirected chordal graph by directing its edges according to one of its "perfect" orderings, and the undirected graph corresponding to a directed chordal graph is always chordal.

A particularly useful result is that given a total ordering there is a unique minimal directed chordal graph, based on the minimal chordal graph fill-in procedure applied to the moral graph. The directed chordal graph is said to be *consistent* with the original influence diagram if the value nodes have no children, and the parents of each decision node include all of the requisite observations and only nodes observed before the decision is made [Nielsen and Jensen, 1999; Shachter, 1999]. When the target order is consistent with the original influence diagram, the fill-in procedure is guaranteed to produce a consistent directed chordal graph.

**Proposition 1** (Minimal Directed Chordal Graph). *Given any influence diagram and a consistent target order, there is a unique minimal directed chordal graph consistent with the order. To construct it from the moral graph, use the order to direct all edges, and visiting each node in reverse order, add a directed arc, if necessary, between each pair of its parents.*

In Figures 1c and 2c we show directed chordal graphs based on the consistent target orders given in the captions. We want to create a tree structure, similar to a junction tree, for the decision circuit backbone, and we use the longest paths in the chordal graphs, comprising the bolded arcs. We can do this based on the following result.

**Theorem 1** (Directed Chordal Graph Properties). *Given a single component directed chordal graph, there is exactly one root node.*

*There is exactly one longest path to each node X from the root and that path contains all of X's ancestors.*

*The arcs on the longest paths form a tree and all of the children of any node X are contiguous with X in the tree.*

*Proof.* There is a path containing all of the ancestors of any node $X$, or two of its ancestors would have a common child but not an arc between them and the graph would not be chordal. That path is the longest path to $X$. There can be only one root node in the graph because each non-root node has exactly one root ancestor. The longest paths form a tree because the ancestors of each node are totally ordered. Finally, if any node $X$ is a parent of $Y$, the node preceding $Y$ in the longest path is either $X$ or a child of $X$, because $X$ and that node are both parents of $Y$ in a chordal graph. (This is equivalent to the "join tree property".) □

### 4.2 Decision circuit backbones

A *decision circuit backbone* is a compact representation for a decision circuit showing the branching as well as sum, product, and max operators. It is compact because it represents a generic path through a symmetric decision circuit. At any time in the construction process, there is a set of variables $\mathbf{W}$ which have been introduced and are still available, and for each $\mathbf{w}$ there is a corresponding operator node in the decision circuit. An uncertainty $X$ is introduced via "$+X|\mathbf{W}$", representing a sum over $x$ for each possible $\mathbf{w}$. Similarly, a decision variable $X$ is introduced via "$\max X|\mathbf{W}$", representing a max over $x$ for each possible $\mathbf{w}$. On the other hand, no variable is introduced via "$*X|\mathbf{W}$", representing a product with $\theta_{x|\mathbf{u}}$ and $\lambda_x$ over $x$ for each possible $\mathbf{w}$, and it is necessary that the entire family of $X$, including $X$, be contained within $\mathbf{W}$. Branches in the circuit represent conditionally independent sub-models to be combined, each with its own contribution to the total value.

To construct a decision circuit backbone from a directed chordal graph:

1. use the structure of the longest path tree, branching when the tree branches, and adding entries for each node in the chordal graph:

    - for uncertainty $X$ with parents $\mathbf{W}$ in the chordal graph, enter "$+X|\mathbf{W}$";
    - for decision $X$ with parents $\mathbf{W}$, enter "$\max X|\mathbf{W}$", followed by "$*X|\mathbf{W}$"; and
    - for value $V$ with parents $\mathbf{W}$, enter "$*V|\mathbf{W}$", and for any terminal branch that doesn't end with a value, append "$*$ zero value"

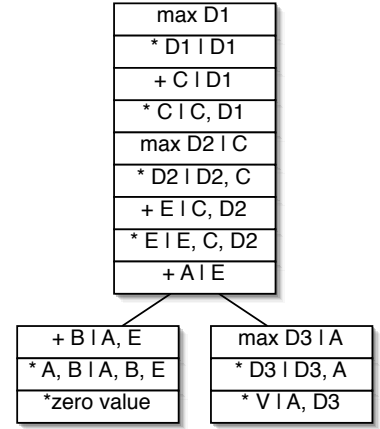

Figure 3: Branching decision circuit backbone for the influence diagram and target order shown in Figure 1.

2. for each chance node $X$, find an entry in the backbone where $\mathbf{W}$ contains the node's family and insert "$*X|\mathbf{W}$" below it. (The moralizing step guaranteed that there will be at least one such entry, and, as a heuristic, we suggest the highest such entry in the backbone.)

The decision circuit backbone corresponding to the chordal graph in Figure 1c is shown in Figure 3, and the backbone corresponding to the chordal graph in Figure 2c is shown in Figure 4. Note that product entries for multiple uncertainties can be combined into one entry, such as "$*A,B|A,B,E$" in Figure 3.

### 4.3 Constructing the decision circuit

In this section we construct the decision circuit from the decision circuit backbone. The backbone organizes all of the information needed for the process.

We describe the construction of the decision circuit starting at the root node with $\mathbf{W} = \{\}$. During the process there will be an operator node created for each possible $\mathbf{w}$. The sum entry "$+X|\mathbf{W}$" and max entry "$\max X|\mathbf{W}$" create a sum (or max) operator node for each possible $\mathbf{w}$ with a child corresponding to each possible $x$. The product entry "$*X|\mathbf{W}$" creates a product operator for each possible $\mathbf{w}$ with three children: the next operator as one child, and $\theta_{x|\mathbf{u}}$ and $\lambda_x$ as the other children (where $\mathbf{u} \subseteq \mathbf{w}$ are the parents of $X$ in the original influence diagram).

For example, the decision circuit for the example in Figure 1 and backbone in Figure 3 corresponds to the following system, equivalent to the one in Section 3:

$$g(\mathbf{e}) = \sum_c \lambda_c \theta_{c|d1} \sum_e \lambda_e \theta_{e|c,d2} \sum_a [g_1(\mathbf{e}_1) g_2(\mathbf{e}_2)]$$

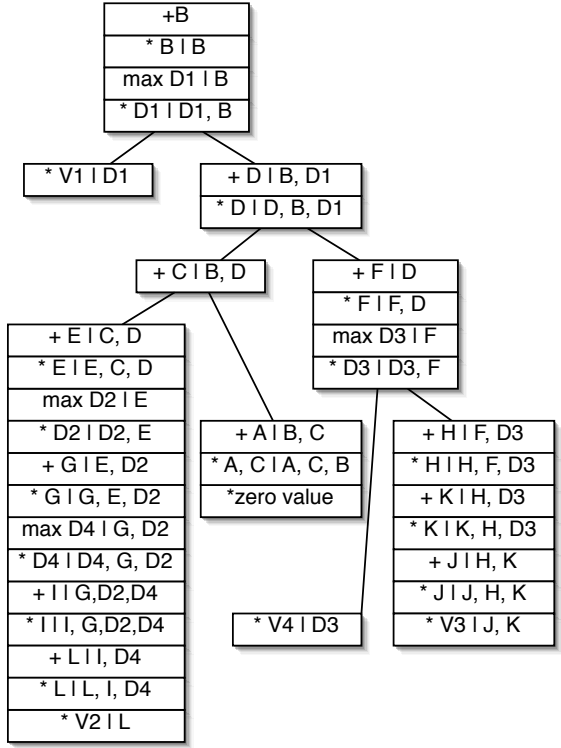

Figure 4: Branching decision circuit backbone for the influence diagram and target order shown in Figure 2.

$$g(\mathbf{e}') = \max_{d1} \lambda_{d1} \sum_c \lambda_c \theta_{c|d1}$$
$$\max_{d2} \lambda_{d2} \sum_e \lambda_e \theta_{e|c,d2}$$
$$\sum_a [g_1(\mathbf{e}'_1)g_2(\mathbf{e}_2) + g_1(\mathbf{e}_1)g_2(\mathbf{e}'_2)]$$
$$g_1(\mathbf{e}_1) = \sum_b \lambda_a \theta_{a|b,e} \lambda_b \theta_b(1)$$
$$g_1(\mathbf{e}'_1) = 0$$
$$g_2(\mathbf{e}_2) = 1$$
$$g_2(\mathbf{e}'_2) = \max_{d3} \lambda_{d3} \lambda_v \theta_{v|a,d3}$$

Given non-responsive evidence $\mathbf{e}$, sweep upward in the constructed decision circuit to compute $g(\mathbf{e})$ and $g(\mathbf{e}')$. For any strategy $s$, $g(\mathbf{e}) = P\{\mathbf{e}|s\} = P\{\mathbf{e}\}$, and

$$g(\mathbf{e}') = \max_s E\{\sum_V \lambda_v V|s, \mathbf{e}\} P\{\mathbf{e}|s\}$$
$$= g(\mathbf{e}) \max_s \sum_V \lambda_v E\{V|s, \mathbf{e}\}.$$

This approach builds on Darwiche [2000, 2003] and Bhattacharjya and Shachter [2007] and leads to the following result. Because the time and space complexity are governed by the size of the decision circuit, we can relate the complexity to the structure of the directed chordal graph determined by the original influence diagram and target variable elimination order. The tree-width of the influence diagram evaluation is exactly the number of nodes in the largest family in the directed chordal graph. (By contrast, the construction of the decision circuit backbone is only on the order of the number of edges in that directed chordal graph.)

**Theorem 2** (Decision Circuit Evaluation). *Given an influence diagram and consistent target variable elimination order, the decision circuit constructed from the directed chordal graph and decision circuit backbone will correctly compute $P\{\mathbf{e}\} = g(\mathbf{e})$ and maximal expected value $g(\mathbf{e}')/g(\mathbf{e})$ for the diagram.*

*The time and space complexity for this computation and for the computation of all of the partial derivatives of both functions with respect to the influence diagram parameters is of the order of $O(ns^t)$, where $t$ is the treewidth (in our case, the size of the largest family in the directed chordal graph), $n$ is the number of variables in the influence diagram, and $s$ is the largest number of states for any variable. Alternatively, the time and space complexity is of the order of $O(nS)$ where $S$ is the largest number of states in any family in the directed chordal graph.*

### 4.4 Comparing Branching and Linear Decision Circuits

In this section, we compare the new branching method presented in this paper with the linear method presented in Bhattacharjya and Shachter [2007], counting the number of arcs in the decision circuits for each approach. We consider three influence diagram examples, the diagrams shown in Figure 1 [Dittmer and Jensen, 1997] and Figure 2 [Jensen, et al, 1994], and the oil wildcatter drilling problem [Raiffa, 1968]. For the first two we construct circuits assuming all variables are binary and also assuming all variables have either three or four states. The target variable elimination orders (shown in the captions in Figure 1 and Figure 2) were chosen to minimize the size of the *linear* decision circuit with the multiple values combined into a single value. The results are shown in Table 1.

Table 1: Comparative size of linear and branching decision circuits for different versions of the influence diagrams shown in Figures 1 and 2, and the oil wildcatter problem in Raiffa [1968].

| Example | Linear | Branching |
|---|---|---|
| Figure 1, binary | 138 | 138 |
| Figure 1, 3 states | 396 | 387 |
| Figure 1, 4 states | 860 | 828 |
| Figure 2, binary | 1,668 | 512 |
| Figure 2, 3 states | 13,755 | 1,812 |
| Figure 2, 4 states | 64,576 | 4,664 |
| Oil Wildcatter | 1,746 | 1,320 |

For the influence diagram shown in Figure 1 with a single value node and only a small opportunity for branching, the overhead of branching cancels out the benefit when there are only two states, and most of the benefit when there are more states. However, in the other two problems where there are multiple additive values, the benefit is much more substantial than the overhead, and that benefit dramatically increases with the number of states.

## 5 DECISION CIRCUITS AND LOCAL STRUCTURE

In this section, we discuss some types of local structure in influence diagrams and how we can exploit it to compile more compact decision circuits.

Influence diagrams naturally represent symmetric decision situations at the graphical level, while any asymmetry is represented at the local level within the distributions. Bielza and Shenoy [1999] compare how different graphical models deal with asymmetry.

It is natural for real world decision problems to exhibit asymmetry, with many zeros in CPTs. To exploit local structure when compiling a decision circuit, it matters whether a particular parameter might change. For example, if we want a particular zero to always stay zero, even during sensitivity analysis, we call it a *hard* zero. The presence of hard zeros and ones allows us to *prune* the decision circuit, eliminating nodes and arcs [Bhattacharjya, 2008]. Situations where we can prune an *intermediate* node (neither a leaf nor a root) or an arc include: a node with a hard 0 child, a product node with a hard 1 child, a node with only one child, and a node whose incoming arcs have all been pruned.

Another way to create more compact decision circuits is with *coalescence*, giving a node multiple parents. Coalescence has usually been hand-crafted for decision trees, as it has been difficult to automate [Bielza and Shenoy, 1999]. Conditional independence is a rich source of coalescence in decision circuits at the global level [Darwiche, 2000], as is the branching we exploit in this paper. When there is no coalescence in the circuit, the circuit resembles a full tree, exponential in the number of variables. In general, the size of the decision circuit is exponential in the size of the largest family in the directed chordal graph, and conditional independence and branching in the global structure can be exploited to limit the circuit size.

Coalescence can be employed at the local level whenever the same symbolic sub-circuit can be re-used [Chavira, 2007]. This can be most easily recognized at the lowest level. For example, if a product node for $x|\mathbf{u}$ is created for $\lambda_x \theta_{x|\mathbf{u}}$, it can be reused whenever $\mathbf{U} \subset \mathbf{W}$ in "$*X|\mathbf{W}$".

## 6 CONCLUSIONS AND FUTURE RESEARCH

In this paper we have shown how significantly smaller and more efficient decision circuits can be built than with the linear vertex elimination method in Bhattacharjya and Shachter [2007]. Given a target variable elimination order, we are able to exploit separate value nodes and conditionally independent subproblems to construct a branching structure over subproblems. We introduce a branching operator and a decision circuit backbone as a compact graphical representation for a symmetric decision circuit. We show how the influence diagram and target order determine a directed chordal graph to guide the construction process. The resulting circuit can be compiled, exploiting the local structure.

The promise of decision circuits is that by compiling the fundamental operations used in junction tree methods exploiting conditional independence and separable values with local operations exploiting asymmetry, we should be able to outperform other exact methods. The advances in this paper help us realize some of those benefits.

Once we have a compiled decision circuit, it can efficiently address a variety of queries in real time, and perform a full sensitivity analysis [Bhattacharjya and Shachter, 2008, 2010]. We are not only able to compute the optimal policy as we manipulate the input parameters, but also the derivative of our objective with respect to all of those parameters. Because the compiled circuit accepts evidence, we can compute quantities such as the value of clairvoyance that can not be determined from the derivatives, taking full advantage of any separable value function in our influence diagram.

We can easily incorporate a few enhancements. First, we can recognize circumstances under which some alternatives might not be available [Smith et al, 1993]. Second, we could introduce product branching in addition to our (additive) branching to allow value functions with nested sums and products as in Tatman and Shachter [1990]. The corresponding operator is just the existing product operator. Finally, instead of computing both $g(\mathbf{e}')$ and $g(\mathbf{e})$ using the same circuit, we could build a simpler arithmetic circuit to compute $g(\mathbf{e})$, reducing compute time, and creating even more opportunities for pruning during compilation.

**Acknowledgements**

We thank the anonymous referees for their suggestions.